\documentclass{article}
\usepackage{spconf,amsmath,graphicx}
\usepackage{amssymb}
\usepackage{multirow}

\title{Intra-Ensemble in Neural Networks}
\name{\textsuperscript{1}Yuan Gao\thanks{Yuan Gao and Zixiang Cai contribute equally.}, \textsuperscript{2}Zixiang Cai\footnotemark[\value{footnote}], \textsuperscript{3}Lei Yu}
\address{\textsuperscript{1}Megvii Inc (Face++), \textsuperscript{2, 3}Beihang University}
\begin{document}
%
\maketitle
\begin{abstract}
Improving model performance is always the key problem in machine learning including deep learning.
However, stand-alone neural networks always suffer from marginal effect when stacking more layers. 
At the same time, ensemble is an useful technique to further enhance model performance.
Nevertheless, training several independent deep neural networks for ensemble costs multiple resources.
If so, is it possible to utilize ensemble in only one neural network?
In this work, we propose \textbf{Intra-Ensemble}, an end-to-end ensemble strategy with stochastic channel recombination operations to train several sub-networks simultaneously within one neural network.
Additional parameter size is marginal since the majority of parameters are mutually shared.
Meanwhile, stochastic channel recombination significantly increases the diversity of sub-networks, which finally enhances ensemble performance.
Extensive experiments and ablation studies prove the applicability of intra-ensemble on various kinds of datasets and network architectures.
\end{abstract}
\section{Introduction}
\label{sec:intro}
Ensemble learning has been proved impactful in traditional machine learning
\cite{breiman2001random, liu1999ensemble, caruana2004ensemble}. 
Concurrently, it is widely applied in deep learning as well. 
It is utilized to enhance final performance on different tasks~\cite{liu2018path}.
Nevertheless, most applications of ensemble in deep learning follow the traditional strategy, simply combining several individually trained neural networks. 
This strategy introduces multiple extra parameters and computational resources, which is extravagant for most practical applications.
 \begin{figure}[ht]
  \centering
  \includegraphics[width=\linewidth]{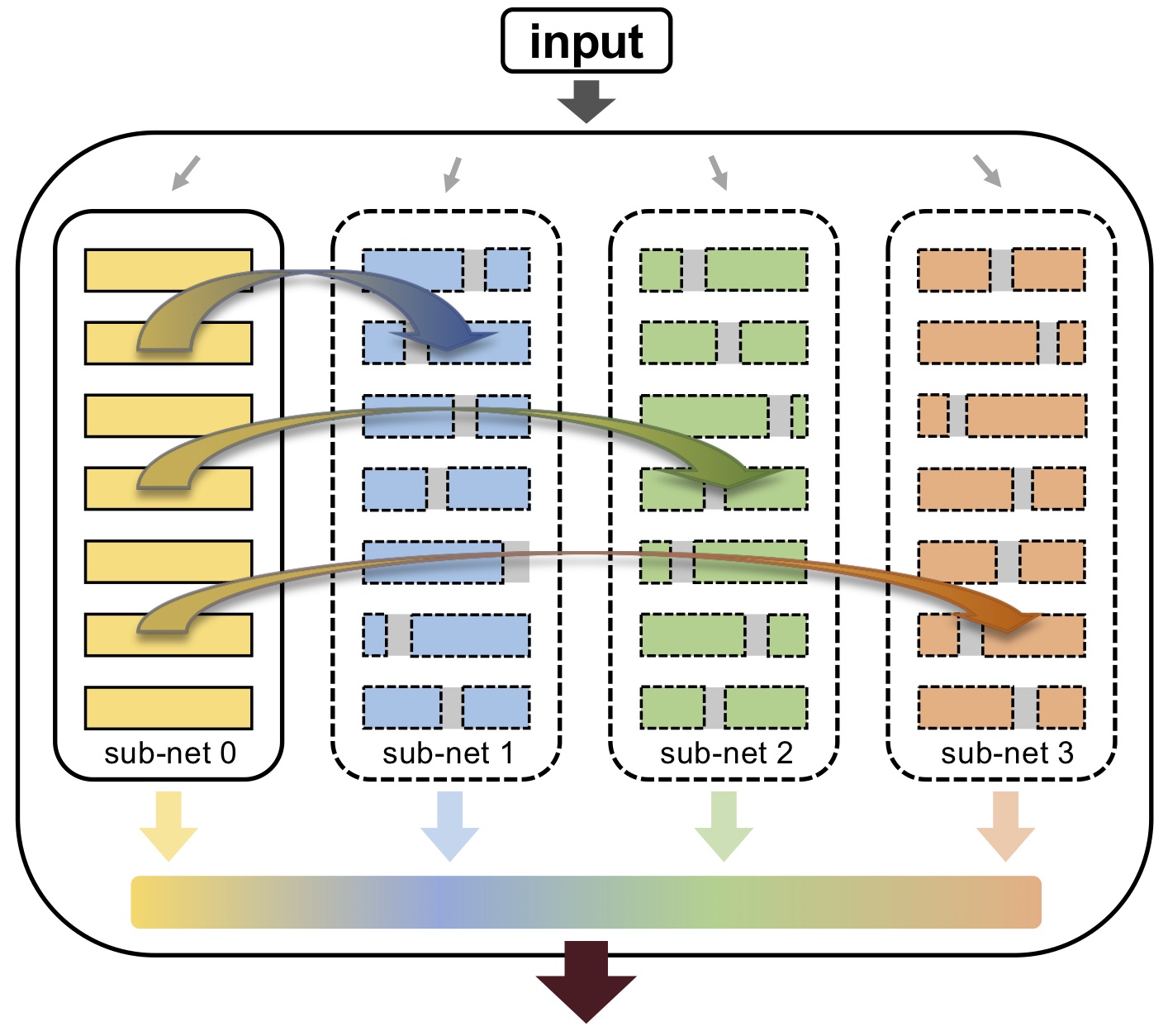} 
  \caption{Intra-Ensemble Network (IENet). Sub-network 0 indicates the original network with all channels and layers. Sub-network 1, 2, 3 share different channels and layers of the original network.} 
  \label{sketch_map}
 \end{figure}
The design of many famous CNN architectures was partly inspired by the idea of ensemble.
The most famous Inception series~\cite{szegedy2017inception} concatenate different branches with various filter types or depths in each layer.
Other architectures like ResNet~\cite{he2016deep} and DenseNet~\cite{huang2017densely} sum or concatenate different features from previous layers.

Our work is mainly inspired by one-shot model~\cite{bender2018understanding} and slimmable neural networks~\cite{yu2018slimmable}. 
One-shot model proposes to train a neural network with multiple optional operations in each position with certain drop probability.
Different sub-networks are naturally generated by keeping different operations at each position.
The positive correlations make it possible to estimate the stand-alone model's accuracy using its corresponding sub-network in one-shot model.
However, owing to relatively large search space, there are thousands of underlying combinations to form various sub-networks.
Parameters shared in sub-networks conflict with each other, leading to a serious accuracy decrease compared to corresponding stand-alone networks. 
A general method~\cite{yu2018slimmable} to train a single network executable at different widths by switchable batch normalization(S-BN) has been proposed. 
Different sub-networks mutually share weights so total parameter size is nearly the same as the stand-alone one, except marginal parameters brought in by S-BN.
In our view, S-BN could provide high performance sub-networks, which are fundamental portions for our ensemble technique.

Another important question is, could sub-networks reach similar accuracies with the original network? 
In our experiments, we find that it is possible when the network size is relatively over-parameterized on the training dataset. 
Based on the common sense that deep neural networks are usually over-parameterized, we attempt to make use of this redundancy in deep neural networks by applying intra-ensemble among inner sub-networks.
However, training a switchable network directly will create a series of sub-networks with similar properties, which is obviously harmful to ensemble.
For this reason, the key problem is to train high-accuracy and low-correlation sub-networks to keep their ensemble ability.
Once sub-networks have more uncorrelated outputs to solve different kinds of hard cases, ensemble can take effect and contribute to the final performance. 
To this end, we propose intra-ensemble with stochastic channel recombination operations(a.k.a. IENet) to produce efficient diversity among sub-networks.
The sketch-map in figure~\ref{sketch_map} shows the relation between original network and sub-networks and the conception of intra-ensemble.
With a fixed size of network parameters, we can generate several sub-networks with stochastic channel arrangement.
Based on the properties of high-accuracy and diversity, we implement simple combination strategies on sub-networks rather than picking up the best one described in one-shot method.
We interestingly find that, although sub-networks may suffer a tiny accuracy drop, intra-ensemble result can still surpass the stand-alone neural network with almost same parameter size.

The main contributions of this work are as follows:
\begin{itemize}
    \item We originally propose intra-ensemble with stochastic channel recombination operations to significantly increase the diversity of sub-networks, which enhances network performance with nearly same parameter size.
    \item We prove the applicability of intra-ensemble on various datasets and network architectures through extensive experiments. Our IENets achieve highly competitive results on CIFAR-10, CIFAR-100 and other tasks.
\end{itemize}

\section{Intra-Ensemble in One Neural Network}

\subsection{Train Sub-networks with Multiple Widths}
In this work, parameter redundancy is tactfully utilized to train several sub-networks within one neural network while sharing most of the network weights.
Here we define a list of width ratios $W$ to control the channel number in each layer. 
For any width ratio $w_i \in W$, $w_i$ is the ratio of used channels per layer.
To be specific, giving a layer with $n$ channels, if $w_i=0.8$, the corresponding $i$-$th$ sub-network use $0.8n$ channels of the original network in this layer.
However, naive training with different widths will cause a great performance decline for each sub-network.
To address this issue, we use slimmable networks~\cite{yu2018slimmable} as a reference and apply their idea of switchable batch normalization(S-BN) in our work.
With S-BN, we can easily train one neural network at different widths with a marginal increase of parameter size, while keeping their high performance: Layers with different widths employ independent batch normalization to accumulate their own feature statistics, which ensures stability in both training and inference procedures.
It is different from the original work that we utilize S-BN with several sub-networks using more than 90\% channels in every layer of the network.
Even if the parameter sharing rate is much higher in our case, S-BN can still ensures the sub-networks' performance.


\subsection{Stochastic Channel Recombination}\label{diversity}
Naively trained sub-networks using slimmable methods~\cite{yu2018slimmable} usually converge towards similar results because they share a large part of parameters with each other.
Ensemble demands a set of diverse models to go into effect.
So the crucial point is to increase the diversity among different sub-networks.
Here we introduce intra-ensemble with stochastic channel recombination operations to significantly enhance ensemble performance within one neural network.
In order to reduce the homogeneity among sub-networks, we resolve the problem into creating structural difference among sub-networks.
We propose the following three recombination operations shown in figure \ref{diff_ops} to increase sub-network diversity on width.
Our operations are mainly implemented on rearranging channel indexing.
Suppose we have a layer with $c$ channels in total, with a sub-network containing $n(0 < n \leq c)$ channels in this layer. 
The operations are described as follows:

\begin{figure*}[ht]
 \centering
 \includegraphics[width=\linewidth]{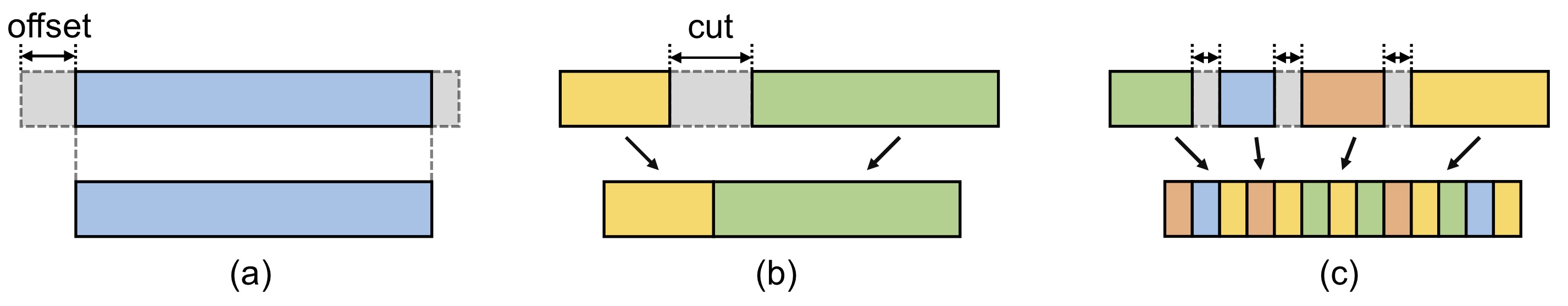}
 \caption{Different stochastic channel recombination operations. (a) random offset. (b) random cut. (c) shuffle channel.} 
 \label{diff_ops}
\end{figure*}

{\bf Random cut (RC)} Random cut means randomly cutting out a continuous part of channels from all channels for sub-network.
If we cut out $p$ percent channels with a cut index $t$ which is randomly generated, we will block out channels with index  $[t, t+1,...,t+pc) $, channels in $[0,...,t)$ and $[t+pc,c)$ are remained for the forward and backward propagation of corresponding sub-network.
So the channels used for a certain layer can be described as:
$$ \pmb w_{used} = \left [\pmb w_i \right]_{i \in \left[ 0,...,t,t + pc,...c \right)} $$ 
with t randomly generated from $[0,...,c-pc)$ and $w_i$ i-th channels in this layer.

{\bf Random offset (RO)} Random offset operation sets offsets for sub-network channels, instead of simply choosing channels starting from the head.
If one sub-net layer use $p$ percent of total channels $c$ with offsets $t$, its channel index list will be $[t,t+1,t+2,...,t+pc)$. 
The constraint of offset $t$ is $0\leq t < c-pc$.
So the channels used for a certain layer can be described as:
$$ \pmb w_{used} = \left [\pmb w_i\right ]_{i\in \left [ pc,...c\right ) } $$ 
with t randomly generated from $[0,...,c-pc)$ and $w_i$ i-th channels in this layer.

{\bf Shuffle channel (SC)} Shuffle channel is inspired by ShuffleNet~\cite{ma2018shufflenet}.
We randomly choose different lists of channel indexes and concatenate the features with shuffled order for sub-network layers.
With shuffle channel operation, different sub-network will use different channels with various order. In this way, diversity can be greatly enhanced.

{\bf Sub-network similarity} Here we propose a simple metric called similarity noting $\mathcal{S}$.
Given test dataset with $N$ test images, if there are $K$ images with same outputs from all sub-networks, the similarity $\mathcal{S}$ is 
$$\mathcal{S}=K/N$$
More similar outputs of sub-networks lead to a higher $\mathcal{S}$ score.
It is a empirically-dependent approach to optimize the balance between similarity and accuracy. 
Adequate experiments are carried out to find the best operation for IENet.

\subsection{Ensemble Strategy}
Since we already have several high-performance and diverse sub-networks, the next step is to decide how to combine the outputs properly.
We simply apply two basic combination strategies of ensemble learning as follows:
Consider the $C$ classes softmax outputs of $N$ sub-networks $\{\pmb o_i \}_{i={1,2,...,N}}$, $\pmb o_i = [x_{i1},x_{i2},...,x_{ic}] \in \mathbb{R}^C $.

{\bf Averaging} Averaging means the probability for each category will be the average of all the sub-networks. The final output is the mean value over softmax outputs.
$$ \pmb o_{avg} = \frac{1}{N} \sum_{i=0}^N \pmb o_i$$

{\bf Stacking} We can also set a small stacking network, which each class will have its privately-owned weights for training and inference, and there is no interactive information among different classes to reduce parameter size to $NC$:
$$ \pmb o_{stacking} =  diag(\pmb W \cdot \pmb O)$$ 
$$ \pmb W \in \mathbb{R}^{C \times N}, \pmb O \in \mathbb{R}^{N \times C} $$

Averaging is a simple and effective parameter-free methods and does not need extra training. 
While benefit by supervised information, stacking can achieve slightly better accuracy in most cases, with marginal parameters added.
Based on practical experience, we mainly choose and report the performance of intra-ensemble with random cut and stacking.


\section{Experiments and Results}


\begin{table*}[ht]
\centering
\begin{tabular}{lcccccc}
    \hline
    Datasets & Classes & Baseline acc. & IENet acc. & Acc. $\Delta $ & Param $\Delta $\\
    \hline
    SVHN~\cite{netzer2011reading} &  10  & 97.59 & {\bf 97.93} & +0.34 & +0.09M\\
    Fashion-MNIST~\cite{xiao2017fashion} & 10 & 96.05 & {\bf 96.43} & +0.38 & +0.09M\\
    Mini-ImageNet~\cite{vinyals2016matching} & 100    & 72.43& {\bf 74.71} & +2.28 & +0.11M\\
    Oxford-IIIT Pets~\cite{em2017incorporating} & 37 & 92.86 & {\bf 94.91}& +2.05  & +0.11M\\
    FGVC Aircraft(Fam.)~\cite{maji2013fine} & 70 & 82.60 &{\bf 87.22}& +4.62 & +0.11M\\
    FGVC Aircraft(Var.)~\cite{maji2013fine} & 102 & 75.34 & {\bf 80.91 }& +5.57 & +0.11M\\
    Caltech-101~\cite{fei2007learning} & 101 & 84.50 & {\bf 87.65} & +3.15 & +0.11M\\
    Food-101~\cite{bossard14} & 101 & 82.27 & {\bf 85.00} & +2.73 &  +0.11M\\
    \hline
\end{tabular}
{
    \caption{Results on other datasets. All experiments are implemented using 4 sub-networks IENet with random cut(RC) and stacking. {\bf Acc. $\Delta$} means the accuracy difference between IENet and the stand-alone baseline network. {\bf Param $\Delta$} means the parameters added with intra-ensemble.}
     \label{other_results}
}
\end{table*}

\subsection{Results on Different Datasets}

Our training setup follows the CIFAR-10 implementation of DARTS~\cite{liu2019darts}.
Data augmentation is exactly the same as DARTS with cutout~\cite{devries2017improved}.
Moreover, we do not add any auxiliary head to assist training.

{\bf CIFAR-10 and CIFAR-100} The typical configuration for CIFAR-10 experiments is: 4 sub-networks with random cut(RC) and a width ratio list $\left[0.9, 0.9, 0.9, 1.0\right]$. It reaches 2.61\% test error, using only 2.66M parameters. 
With less parameters, it outperforms most neural architecture searched and manually designed models.
Moreover, we have a wider version with 4.22M reaching 2.47\% test error and a narrower version with 1.63M reaching 2.91\%, which proves the high scalability of intra-ensemble. 
The comparison with our models and others can be found in table~\ref{CIFAR-10}.
All the models with intra-ensemble have great improvement in accuracy while introducing little extra parameters.
By the way, the 2.57M stand-alone baseline network architecture is a simply modified MobileNet V2 1.0x with slightly wider channels in each layer which attains 3.10\% test error.
We directly apply CIFAR-10 configuration to CIFAR-100, except the output classes number.
Though class number increases from 10 to 100 and bad cases become more complex, our intra-ensemble still have significant effects on improving performance.
The 4 sub-networks IENet with 2.78M parameters have a 1.64\% marginal gain compared to single model with similar parameter size, as in table~\ref{CIFAR-100 Results}.
Surprisingly, the 5 sub-networks IENet with 2.82M parameters have a 2.62\% marginal gain. We conjecture it is due to the relatively low top-1 accuracy compared with CIFAR-10. 
So intra-ensemble has more potentiality for improvement.
\begin{table}[ht]
    \centering
    \begin{tabular}{lccc}
    \hline
    Method & Param. & $\mathcal{S}$ & Err.\\
    \hline
    DenseNet-BC~\cite{huang2017densely} & 25.6M &- & 3.46 \\
    NASNet-A~\cite{zoph2018learning} & 3.3M & - & 2.65\\
    AmoebaNet-A~\cite{real2018regularized} & 3.2M & - &3.34\\
    Darts~\cite{liu2019darts} & 3.3M &- & 2.76 \\
    ILRAS~\cite{guo2018irlas} & 3.91M &- & 2.60 \\
    \hline
    our stand-alone & 2.57M & - & 3.10 \\ 
    \hline
    IENet (4 sub-nets, SC, S) & 2.66M & 0.838 & 2.86 \\
    IENet (4 sub-nets, RO, S) & 2.66M & 0.845  &2.77 \\
    IENet (4 sub-nets, RC, S) & 2.66M & 0.857 &2.61 \\
    \hline
    IENet (4 sub-nets, RC, S) & 1.63M & - &2.91 \\
    IENet (4 sub-nets, RC, A) & 2.66M & - &2.66 \\
    IENet (4 sub-nets, RC, S) & 4.22M & - &2.47 \\
    IENet (5 sub-nets, RC, S) & 2.70M & - &2.66 \\

    \hline
\end{tabular}
{
    \caption{Comparison of test error on CIFAR-10. In the annotations, `S' means Stacking; `A' means Averaging.}
     \label{CIFAR-10}
}
\end{table}

\begin{table}[ht]
\centering
\begin{tabular}{lccc}
    \hline
    Method & Param & $\mathcal{S}$ & Err.\\
    \hline
    DenseNet-BC~\cite{huang2017densely} & 25.6M & - &17.18 \\
    SMASHv2~\cite{brock2017smash} &16M &- & 20.6 \\
    ENAS~\cite{pham2018efficient} & 4.6M &- & 17.27\\
    PNAS~\cite{liu2018progressive} & 3.2M &- & 17.63 \\
    AmoebaNet-B~\cite{real2018regularized} & 34.9M & - &15.80\\
    \hline
    our stand-alone & 2.71M &- & 18.66 \\
    \hline
    IENet (4 sub-nets, SC, S) & 2.78M &0.653 & 18.04 \\
    IENet (4 sub-nets, RO, S) & 2.78M & 0.685 &17.10 \\
    IENet (4 sub-nets, RC, S) & 2.78M & 0.759 & 17.02 \\
    \hline
    IENet (5 sub-nets, RC, A) & 2.82M & - & 16.10 \\
    IENet (5 sub-nets, RC, S) & 2.82M & - & 16.04 \\
    \hline
\end{tabular}
{
    \caption{Comparison of test error on CIFAR-100. In the annotations, `S' means Stacking; `A' means Averaging.}
     \label{CIFAR-100 Results}
}
\end{table}

{\bf Other datasets} We additionally carry out sufficient experiments on different kinds of datasets to prove the solidity of our method as shown in table~\ref{other_results}.
Simple training skills and fewer training epochs are applied in order to quickly verify the effectiveness of intra-ensemble on these datasets.
All the extra experiments are carried out on 4 sub-networks IENet using random cut(RC) and stacking.
While different datasets have various image sizes and class numbers, intra-ensemble always gains performance improvement on both simple and complex tasks.
The results in the table demonstrate the general applicability of intra-ensemble to various types of data.

{\bf Similarity analysis} A good trade-off between accuracy and similarity should be carefully considered while using different stochastic channel operations.As shown in table~\ref{CIFAR-10} and \ref{CIFAR-100 Results}, these operations significantly decrease sub-network similarity $\mathcal{S}$.
As we can see, intra-ensemble with RC balances the best accuracy-similarity trade-off comparing to RC and RO. Although having relatively lower similarity, the intra-ensemble results using RO and SC can not surpass results using RC because of the worse accuracy drop of sub-networks. We conjecture that it is because when $\mathcal{S}$ gets lower, there are more ``conflicts'' among the weights-shared sub-networks.
Moreover, the similarity $\mathcal{S}$
on CIFAR-100 is much lower than that in CIFAR-10.
Correspondingly, intra-ensemble result has more improvement on CIFAR-100.
We deduce that the complexity and variety of CIFAR-100 lead to greater difference among sub-networks, thus ensemble performance is better enhanced.
Besides, in table~\ref{other_results}, the datasets with more classes also have larger improvement with intra-ensemble.
Presumably, intra-ensemble works better when dealing with more complex tasks which have more image classes.


\section{Conclusion}
We have introduced Intra-Ensemble, a novel strategy which combines several diversified sub-networks within one neural network to enhance final performance.
The stochastic channel recombination operations ensure high-accuracy and diversity.
With marginal parameters added, intra-ensemble achieves competitive results compared with other methods on classification tasks.
Extensive experiments show that our method is effective on various kinds of architectures and datasets.
Besides, as multi-core computing power is more and more widespread, model parallelism will become more easily and sub-networks can be simultaneously trained to save training time. 
In this work, we only carry out experiments on classification tasks. 
But we believe IENet could enhance the performance of other CNN architecture and be applied to other tasks using CNN(e.g. object detection, etc.) to improve model performance.
We will work on it to maximize the utilization of intra-ensemble method.




\bibliographystyle{IEEEbib}
\bibliography{egbib}

\end{document}